\documentclass[letterpaper, 10 pt, journal, twoside]{IEEEtran}
\usepackage{graphics} 
\usepackage{epsfig} 
\usepackage{mathptmx} 
\usepackage{times} 
\usepackage{amsmath} 
\usepackage{amssymb}  
\usepackage{booktabs}
\usepackage{multirow}
\usepackage{multicol}
\usepackage{graphicx}
\usepackage{subfigure}
\usepackage{color}

\usepackage[colorlinks,
            linkcolor=red,
            anchorcolor=blue,
            citecolor=green]{hyperref}

\ifCLASSINFOpdf
\else
\fi
\begin{document}
%
\title{You Only Look Bottom-Up for Monocular 3D Object Detection}
%
%
%

\author{Kaixin Xiong$^{1*}$, Dingyuan Zhang$^{1*}$, Dingkang Liang$^{1}$, Zhe Liu$^{1}$,
\\Hongcheng Yang$^{1}$, Wondimu Dikubab$^{1, 2}$, Jianwei Cheng$^{2}$, Xiang Bai$^{1\dag}$

\thanks{Manuscript received: April, 5, 2023; Revised July, 18, 2023; Accepted August, 25, 2023.}
\thanks{This paper was recommended for publication by Editor Cesar Cadena Lerma upon evaluation of the Associate Editor and Reviewers' comments.
This work was supported by the National Science Fund for Distinguished Young Scholars of China (Grant No.62225603).
} 
\thanks{* Equal contribution}
\thanks{\dag  Corresponding author, \tt\small xbai@hust.edu.cn}
\thanks{$^{1}$Kaixin Xiong, Dingyuan Zhang, Dingkang Liang, Zhe Liu, Hongcheng Yang and Xiang Bai are with Huazhong University of Science and Technology, China. \tt\small \{kaixinxiong, dyzhang233, dkliang, zheliu1994, hcyang, xbai\}@hust.edu.cn}%
\thanks{$^{2}$Wondimu Dikubab and Jianwei Cheng are with JIMU Intelligent Technology Co., Ltd, China. \tt\small wondiyeaby@gmail.com, cjw@jmadas.com}%

\thanks{Digital Object Identifier (DOI): see top of this page.}
}
%
%

\markboth{IEEE Robotics and Automation Letters. Preprint Version. Accepted August, 2023}
{Xiong \MakeLowercase{\textit{et al.}}: You Only Look Bottom-Up for Monocular 3D Object Detection} 

%



\maketitle

\begin{abstract}
Monocular 3D Object Detection is an essential task for autonomous driving. Meanwhile, accurate 3D object detection from pure images is very challenging due to the loss of depth information. Most existing image-based methods infer objects' location in 3D space based on their 2D sizes on the image plane, which usually ignores the intrinsic position clues from images, leading to unsatisfactory performances. Motivated by the fact that humans could leverage the bottom-up positional clues to locate objects in 3D space from a single image, in this paper, we explore the position modeling from the image feature column and propose a new method named You Only Look Bottum-Up (YOLOBU). Specifically, our YOLOBU leverages Column-based Cross Attention to determine how much a pixel contributes to pixels above it. Next, the Row-based Reverse Cumulative Sum (RRCS) is introduced to build the connections of pixels in the bottom-up direction. Our YOLOBU fully explores the position clues for monocular 3D detection via building the relationship of pixels from the bottom-up way. Extensive experiments on the KITTI dataset demonstrate the effectiveness and superiority of our method.
\end{abstract}


%
\IEEEpeerreviewmaketitle

\section{Introduction}
%
%
%
%
\label{intro}
\IEEEPARstart{M}{onocular}
 3D Object Detection has received increasing attention since it provides a low-cost solution for autonomous driving~\cite{chen2016monocular, liu2021yolostereo3d, jung2022fast, kumar2020fisheyedistancenet}. This task aims to estimate objects’ localization, orientation, and dimensions from a single RGB image. The main challenge of this task comes from irreversible depth information lost in the projection process, and recovering the depth information from 2D to 3D is an ill-posed problem.

\begin{figure}[t]
\centering
\includegraphics[width=1\linewidth]{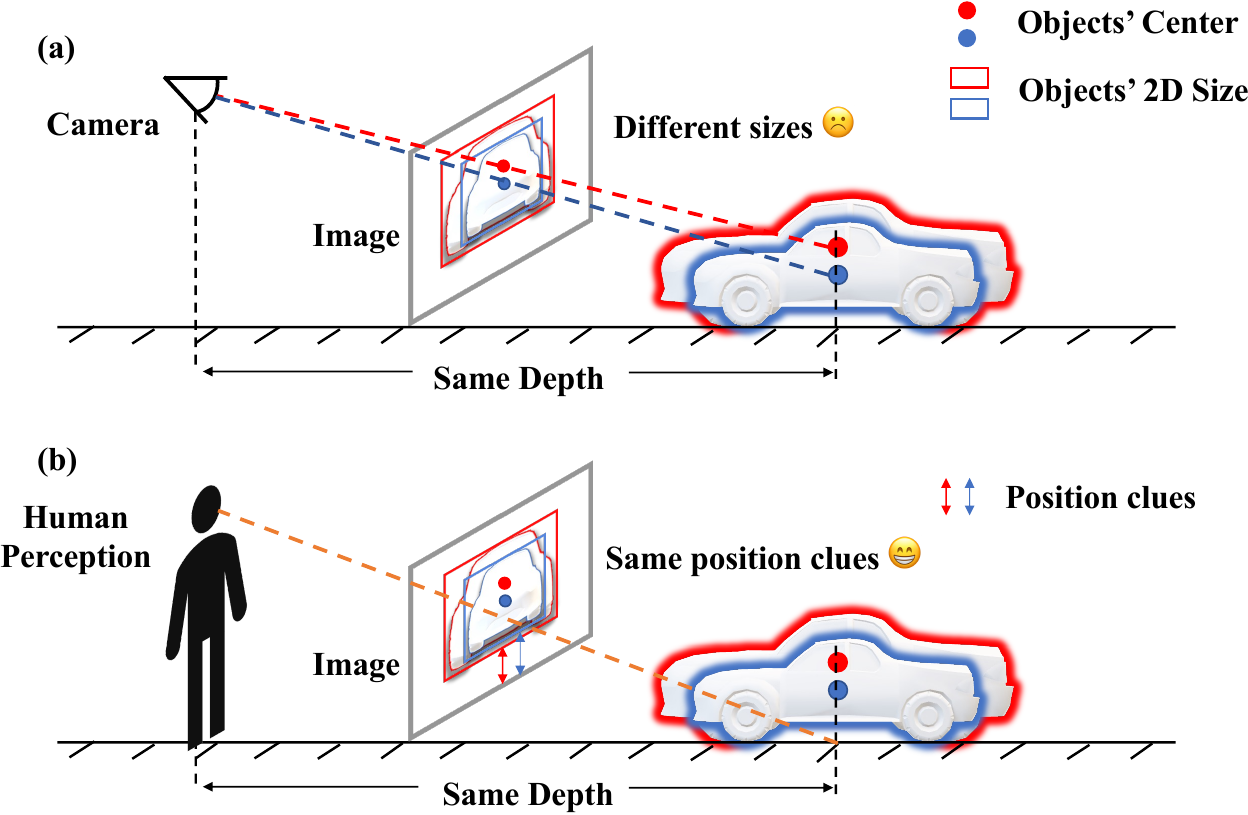}
\caption{Two cars with different dimensions and the same depth and appearance, the corresponding 2D bounding boxes on the image have different sizes. (a) For monocular 3D detectors, ambiguity would occur with only 2D size information; (b) For human perception from monocular images, there is no ambiguity guided by position information.}
\centering
\label{fig:introduction}
\vspace{-10pt}
\end{figure}

Most image-based monocular 3D object detection methods ~\cite{brazil2019m3d, liu2020smoke, lian2021geometry} infer objects’ 3D location mainly via objects’ 2D sizes on the image plane. However, these methods usually fail when facing scale ambiguity problems since only the size clues are leveraged. We present a case on the right of Fig.~\ref{fig:introduction}(a), where two cars are different in 3D dimensions but share the same depth and appearance information. The contradiction comes from detectors should predict the same depth for these two cars while they have different 2D sizes on the image plane.
In other words, such ambiguity could confuse the detector and lead to sub-optimal detection results. Several methods~\cite{liu2021ground, qin2022monoground} have attempted to leverage the ground plane as prior information to solve this problem. However, these methods heavily rely on the strong assumption that the ground plane is flat (i.e., the ground plane is smooth and parallel to the bottom of the ego car),  
which inevitably induces noises when going from horizontal to uphill or downhill. Naturally, digging out the potential 3D prior clues from monocular images is an intuitive and effective solution for avoiding such problems. Indeed, this is consistent with rational behavior when humans locate objects from a single image. 

We observe that humans can easily identify such cases and locate objects accurately with the help of position clues. As shown in Fig.~\ref{fig:introduction}(b), position clues from the earthing contact points of instances to the bottom of the image plane could be treated as explicit references for locating objects, which eliminates the size ambiguity. Inspired by such rational human behaviors, we argue that not only is size information essential for image-based 3D detection, but also the bottom-up positional information on the image plane is indispensable as extra information compensation. 

Note that two prerequisites should be met when utilizing above mentioned position information: 1) the pixels closer to the bottom of the image represent smaller depth; 2) the depth from the bottom of the image to target instances is monotonically increasing. Fortunately, these two assumptions could be easily satisfied in \textit{most} autonomous driving scenes since the camera is mounted on the top of the ego car and the ego car is driven on the drivable areas~\cite{frickenstein2020binary}. 

In this paper, we propose a novel method named You Only Look Bottom-Up (YOLOBU), which naturally and succinctly establishes the association of pixels on the image plane in the bottom-up direction. 
Since not all pixels within the same column and among different columns contribute to the localization of objects above them equally, we introduce the Column-based Cross Attention (CCA) to model such a relationship, which can also help suppress the noise by assigning different attention weights to each pixel when the assumption fails in some conditions (e.g., a large vehicle right in front of your car. The top back edge of that vehicle would have a lower depth than the road straight underneath). Different from the widely used cross-attention in Object Detection~\cite{carion2020end, zhu2020deformable}, our CCA calculates the attention weight based on each image feature column. 
Following CCA, our proposed Row-based Reverse Cumulative Sum (RRCS) is conducted to build the connections of pixels within each column in the bottom-up direction. Benefiting from CCA and RRCS, our YOLOBU can be better aware of the relationship of pixels with different positions on the image plane.

Extensive experiments are carried out on the KITTI dataset to demonstrate the effectiveness of the proposed YOLOBU. In particular, our YOLOBU achieves state-of-the-art for the car category and highly competitive performance for the cyclist/pedestrian category.

In summary, the main contributions of this paper can be summarized in two folds: 1) We provide a rethinking of the position modeling for solving the size ambiguity problem in monocular 3D detection. We point out that the bottom-up position clue is an overlooked design factor for monocular 3D detectors. 2) We propose a new method YOLOBU, a plug-and-play approach that effectively leverages the position clues from the image and can be generalized for other applications.

%

\begin{figure*}[h]
\vspace{10pt}
\centering
\includegraphics[width=0.95\linewidth]{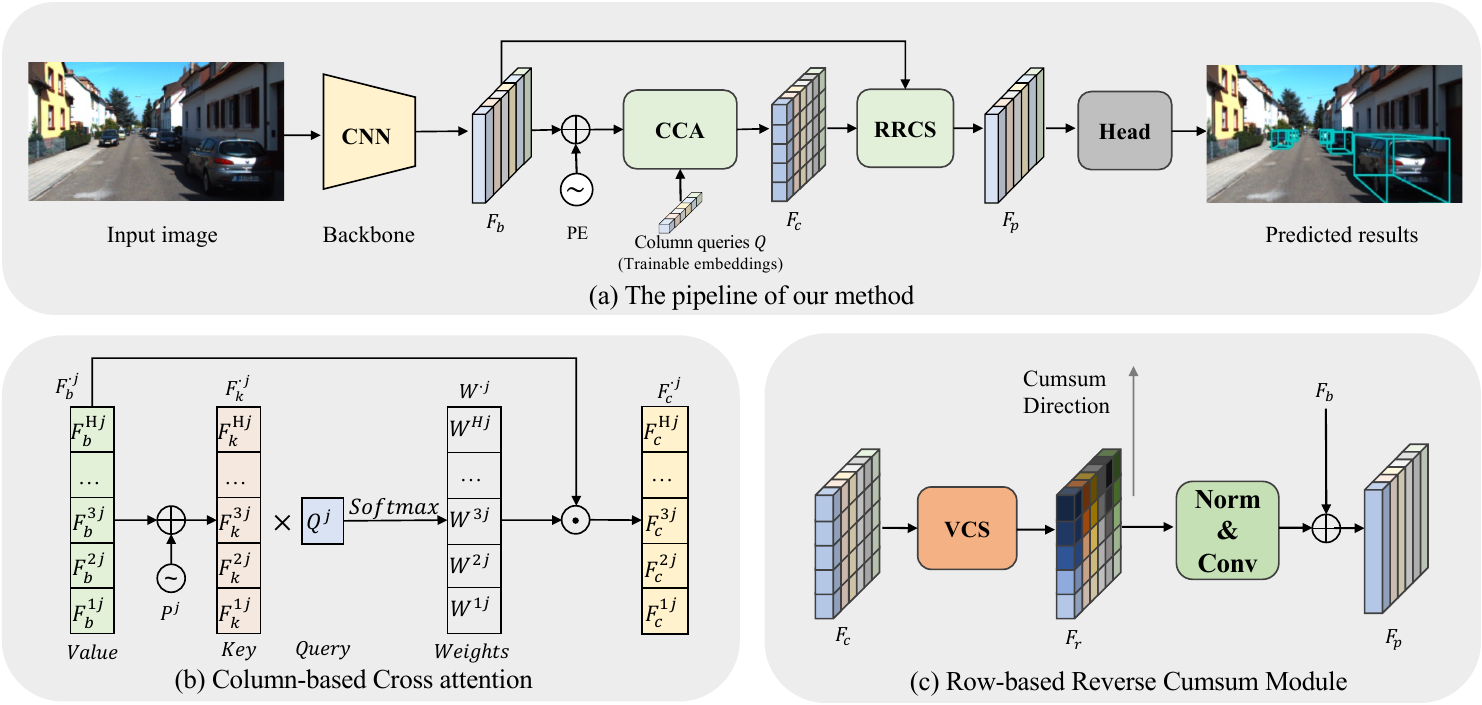}
\caption{\textbf{(a)} The pipeline of the proposed YOLOBU, which consists of a backbone network, our proposed Column-based Cross Attention (CCA) and Row-based Reverse Cumulative Sum (RRCS), and head branches.
\textbf{(b)} The structure of the proposed Column-based cross attention (CCA)  is illustrated on the left of dotted line. And the adjacency matrix is demonstrated on the right of dotted line, where gray Grids denote no connection between those nodes. 
\textbf{(c)} Structure of the proposed Row-based Reverse Cumulative Sum (RRCS). }
\centering
\label{fig:pipeline}
\vspace{-10pt}
\end{figure*}

\section{Related Work}
\subsection{3D Detection with LiDAR}
There are plenty of works that conduct 3D object detection with LiDAR inputs. Many works~\cite{zhou2018voxelnet, yan2018second, lang2019pointpillars,shi2019pointrcnn, liu2020tanet} focus on the pure LiDAR-based 3D object detection with fully supervision. Some other methods~\cite{li2023dds3d, 2023vitwss3d, zhang2023sam3d} explore the semi and weakly supervised learning or even zero-shot learning to reduce the costs of labeling. To take advantage of the complementary between images and point clouds, some methods~\cite{liu2022epnet++, liu2023bevfusion, liang2022bevfusion} propose different strategies for better fusion. Despite their remarkable performance, the dependence on expensive LiDAR sensors hinders the wide use of these methods. Our method only needs a single cheap camera, making it easier for practical usage.

\subsection{Monocular 3D Detection}

Monocular 3D Object Detection methods can be roughly grouped into two categories: depth-guided methods and pure image-based methods.

Depth-guided methods need extra data sources such as point clouds and depth images in the training phase. These methods mainly use the extra sub-network to estimate pixel-level depth under the supervision of the sparse projected point cloud or even dense depth images, which introduces extra labeling efforts. In particular, Pseudo-LiDAR~\cite{wang2019pseudo} converts pixels to pseudo point clouds and then feeds them into a LiDAR-based detector~\cite{shi2019pointrcnn, liu2020tanet, deng2021voxel}. Besides, there are also some methods~\cite{bao2019monofenet, ding2020learning, jing2022depth, xie2021panet} that fuse depth feature and image feature together to improve the performance while ~\cite{ye2020monocular, chong2021monodistill, zhou2022sgm3d} learn 3D prior information via knowledge distillation. Some methods~\cite{philion2020lift, reading2021categorical} conduct view transform with estimated depth distribution. DD3D~\cite{park2021pseudo} claims that pre-training paradigms could replace the pseudo-lidar paradigm. These methods benefit from precise depth information provided by point clouds, which could be considered as using another type of ground truth as extra supervision. 

Pure image-based methods only need RGB images and calibrated information in the whole inference pipeline. These methods mainly learn depth information from objects' apparent size and geometry constraints provided by eight keypoints or pin-hole model~\cite{brazil2019m3d, li2019gs3d, li2020rtm3d, li2022densely, lian2021geometry, liu2022learning, xiong2023cape}. For example, M3D-RPN~\cite{brazil2019m3d} proposes 2D-3D anchor and depth-wise convolution to detect 3D objects. RTM3D~\cite{li2020rtm3d} predicts nine keypoints of 3D proposals and optimizes the proposals via a re-projection cost function. MonoDLE~\cite{ma2021delving} proposes three strategies to alleviate problems caused by localization errors. GUPNet~\cite{lu2021GUPNet} proposes a geometry uncertainty projection method for better localization. MVM3Det~\cite{haoran2021mvm3det} proposes feature orthogonal transformation to estimate the position. TIM~\cite{saha2022translating} conducts monotonic attention to translate images into bird's-eye-view maps. DID-M3D~\cite{peng2022did} decouples the object's depth into visual depth and attribute depth. These methods show great promise with their simple structures and competitive performances.

\subsection{Position Clues for Monocular 3D Detection}
Several methods have exploited the position clues provided by the ground plane. Mono3D~\cite{chen2016monocular} leverages the ground plane to filter redundant proposals. GAC~\cite{liu2021ground} proposes ground-aware convolution that extracts depth priors from the ground plane hypothesis. DeepLine~\cite{liu2021deep} extracts and encodes line segmentation features into image features via Hough transform. MonoGround~\cite{qin2022monoground} generates dense depth supervision from the ground plane of ground truth 3D bounding boxes. These methods could alleviate the size ambiguity problem to some extent. However, they do not fully leverage the positional and contextual information of pixels, which plays an important role in monocular 3D detection. Compared with the above methods, our YOLOBU only requires a weaker assumption that the depth of the ground plane is monotonically increasing from the bottom of the image to target instances, which could be satisfied more easily. Meanwhile, our method doesn't rely on explicit geometry structures in the scene, such as lines or planes. We explicitly build the relationship of pixels in the bottom-up direction, which could serve as a position clue for reasoning objects' location in 3D space.

\subsection{2D-3D Transform for Image-based Perception}
The way to transform 2D image features into features representing 3D space is crucial for image-based perception. It can be roughly categorized into inverse perspective mapping (IPM)-based and depth estimation-based. IPM first assumes a flat plane, and then it maps all pixels from a given viewpoint onto this flat plane through homography projection, and many methods~\cite{kim2019vope},~\cite{palazzi2017ipm_14},~\cite{zhu2021trafcam3d},~\cite{reiher2020cam2bev} use IPM for obstacles detection or lane detection task. Depth estimation-based methods~\cite{reading2021categorical},~\cite{philion2020lift},~\cite{wang2019pseudo},~\cite{wang2021fcos3d},~\cite{peng2022did} mainly estimate pixel-level depth or depth distributions to transform the image features into 3D features. Since our method acts as a plug-and-play module and operates on the image feature level, it can be adopted for both types of approaches.

\section{METHOD}
Fig.~\ref{fig:pipeline}(a) shows the pipeline of our YOLOBU, which contains three parts: 
1) A CNN-based backbone used to extract the feature map $F_b$ from images; 
2) The proposed Colum-based Cross Attention (CCA), which conduct cross attentions between image features and the column-wise queries $Q$ to produce attention weights for each pixel in each column and then get the re-weighted feature $F_c$.
3) The proposed Row-based Reverse Cumulative Sum(RRCS), which effectively makes pixels perceive the scenes below them and outputs the position aware feature $F_p$. 
In the next sections, we delve deeper into the key elements of YOLOBU, i.e., CCA and RRCS. For a better explanation, we set the origin of coordinate system in the \emph{bottom left} of the image.

\subsection{Column-based Cross Attention (CCA)}

Structured layout (i.e., road) pixels with different semantics and locations devote differently to object localization.
For example, in each column, the road pixels closer to the top of the image plane represent a larger depth interval, which may contribute more to the localization of far objects and less useful for near objects. This phenomenon indicates that it is infeasible to treat pixels in the same column equally. Furthermore, the depth interval of pixels in different columns varies, inspiring us to pay attention to the differences among different columns. To model such a relationship, we propose the coloumn-based cross attention mechanism termed as CCA, which leverages cross attention to assign different weights to pixels within each column, and uses a single distinct query for each column to represent the differences between columns. 

Concretely, given the input feature map $F_{b} \in \mathbb{R}^{H \times W \times C} $ extracted from backbone, we define a set of learnable embedding queries $Q \in \mathbb{R}^{W \times C}$. Each query corresponds to a column of the feature map separately. 

We illustrate the mechanism of CCA, taking the $j$-th column as an example. As shown in Fig.~\ref{fig:pipeline}(b), the sine-cosine position encoding $P^{\cdot j} \in \mathbb{R}^{H \times C}$ is added to the image feature $F_b^{\cdot j} \in \mathbb{R}^{H \times C}$ to get position encoded feature $F_{k}^{\cdot j} \in \mathbb{R}^{H \times C}$, as below: 

\begin{equation}
\begin{gathered}
F_{k}^{\cdot j} = F_{b}^{\cdot j} + P^{\cdot j}.
\end{gathered}
\end{equation}


Next, we get the attention weight $W^{\cdot j} \in \mathbb{R}^{H \times 1}$ by conducting matrix multiplication attached with a $Softmax$ activation function:

\begin{equation}
\begin{gathered}
W^{\cdot j} = Softmax\left(F_k^{\cdot j} (Q^j)^{\mathrm{T}} \right),
\end{gathered}
\end{equation}

where $Q^j \in \mathbb{R}^{1 \times C}$ is the $j$-th embedding query, and $W^{\cdot j} \in \mathbb{R}^{H \times 1}$ represents the attention weights for the $j^{th}$ column.

We stack the weights $W^{\cdot j}$ and repeat through the channel dimension to get the complete attention weights $W \in \mathbb{R}^{H \times W \times C}$, and finally we get the output feature $F_{c} \in \mathbb{R}^{H \times W \times C}$ by multiplying $W$ with $F_{b}$:
\begin{equation}
\begin{gathered}
F_{c} = W \bigodot F_{b},
\end{gathered}
\end{equation}
where $\bigodot$ denotes Hadamard product.

Note that our neat CCA module builds the relationships within columns and across columns efficiently.
Compared with the original transformer~\cite{vaswani2017attention, wang2018non} that costs $O((H*W)^2 * C)$, the time complexity of our proposed CCA is only $O(H*W*C)$.

\subsection{Row-based Reverse Cumulative Sum (RRCS)}
After re-weighting the pixels within each column, we need to build the connections for pixels in the one-way direction (i.e., bottom-up direction), since the bottom-up positional information is indispensable as extra information compensation and can serve as references for locating objects, as what we argued in Sec.~\ref{intro}. For this purpose, we propose the RRCS, which models the bottom-up relationship between pixels in a row-by-row way.

Specifically, as illustrated in Fig.~\ref{fig:pipeline}(c), we conduct \textbf{v}ertical \textbf{c}umulative \textbf{s}um (VCS) on the output of CCA module $F_{c} \in \mathbb{R}^{H \times W \times C}$ in the bottom-up direction:

\begin{equation}
F_r^{i \cdot} = \sum_{k=1}^{i} F_c^{k \cdot} , 
\end{equation}

where the $F_r^{i \cdot}$ and $F_c^{k \cdot}$ represents the $i$-th row of feature map $F_r$ and $k$-th row of input $F_c$, respectively. The RRCS promises that each pixel could perceive all of the pixels below it, which could be treated as a dense unidirectional connection at the pixel level. Thus, encoded features of structured hierarchical layout pixels (i.e., road pixels) could be propagated to the target objects, which acts as a position-related clue for 3D pose reasoning.

Note that the cumulative sum would change the value magnitude of features, which might be hard for the network to converge. Thus, normalization is needed here to keep values stable,
which is conducted as the Eq.~\ref{eq: norm_by_mask} shows:

\begin{equation}
\hat{F_r^{i\cdot}} = \frac{F_r^{i \cdot}}{i}.
\label{eq: norm_by_mask}
\end{equation}

The feature map $\hat{F_{r}}$ is then encoded by a $1 \times 1$ convolution layer $\varphi$ and added with the feature $F_{b}$ to formulate the position aware feature $F_{p} \in \mathbb{R}^{H \times W \times C} $:

\begin{equation}
F_{p} = F_{b} + \varphi(\hat{F_{r}}).
\end{equation}

\subsection{Training Objective}
Our method predicts 3D proposals based on the feature map $F_{p}$, which is attached to the detection head. Our detection heads are composed of three branches: classification branch, 2D regression branch and 3D regression branch. Optimizing 2D parameters $b_{2D}$ and 3D parameters $b_{3D}$ simultaneously helps the network learn geometrical relationships between the image and world space.

We follow~\cite{ma2021delving} as for the training loss. Suppose that $y = (c, b_{2D}, b_{3D})$ and $\hat{y} = (\hat{c}, \hat{b_{2D}}, \hat{b_{3D}})$ denote the set of ground truths and predictions respectively, where $c$ denotes the classification score.
The training objective is summarized as Eq.~\ref{eq:total_loss}.
\begin{equation}
    L = L_{cls}(c, \hat{c}) + L_{2D}(b_{2D}, \hat{b_{2D}}) + L_{3D}(b_{3D}, \hat{b_{3D}})   
\label{eq:total_loss}
\end{equation}

Here $L$ is the overall loss, while $L_{cls}$, $L_{2D}$ and $L_{3D}$ represents classification loss, 2D detection loss and 3D detection loss respectively. 

\begin{table*}[t]
\vspace{10pt}
\small
\setlength{\tabcolsep}{3mm}
\centering
\caption{The performance for \textbf{Car} on the KITTI \emph{\textbf{test}} set. The \textbf{Bold} font indicates the first place in this table. $^*$ denotes that EXTRA DATASET is used.}
\vspace{-10pt}
\label{tab:main_result}

\begin{tabular}{ l c c ccc ccc }
\toprule
\multirow{2}{*}{Methods} &\multirow{2}{*}{Year} &\multirow{2}{*}{Extra Data} &\multicolumn{3}{c}{$AP_{3D|R_{40}}$} &\multicolumn{3}{c}{$AP_{BEV|R_{40}}$}\\
\cmidrule(r){4-9}
& & &Easy &Moderate &Hard &Easy &Moderate &Hard\\

\midrule
D4LCN~\cite{ding2020learning} &CVPR20 &Depth &16.65 &11.72 &9.51  &22.51 &16.02 &12.55 \\
CaDDN~\cite{reading2021categorical} &CVPR21 &LiDAR &19.17 &13.41 &11.46  &27.94 &18.91 &17.19 \\
AutoShape~\cite{liu2021autoshape} &ICCV21  &CAD Models &22.47 &14.17 &11.36  &30.66 &20.08 &15.95\\
DD3D~\cite{park2021pseudo} &ICCV21 &Depth$^*$ &23.22 &\textbf{16.34} &\textbf{14.20} &30.98 &22.56 &\textbf{20.03}\\
SGM3D~\cite{zhou2022sgm3d} &RA-L22 &Stereo &22.46 &14.65 &12.97 &31.49 &21.37 &18.43\\
MonoDistill~\cite{chong2021monodistill} &ICLR22 &LiDAR &22.97 &16.03 &13.60 &31.87 &22.59 &19.72\\
DID-M3D~\cite{peng2022did} &ECCV22 &LiDAR &\textbf{24.40}
&16.29 &13.75 &\textbf{32.95} &\textbf{22.76} &19.83\\
\midrule
M3D-RPN~\cite{brazil2019m3d} &ICCV19  &None &14.76 &9.71 &7.42 &21.02 &13.67 &10.23 \\
MonoPair~\cite{chen2020monopair} &CVPR20 &None &13.04 &9.99 &8.65 &19.28 &14.83 &12.89\\
MonoDLE~\cite{ma2021delving} &CVPR21 &None &17.23 &12.26 &10.29 &24.79 &18.89 &16.00\\
MonoEF~\cite{zhou2021monoef} &CVPR21 &None &21.29 &13.87 &11.71  &29.03 &19.70 &17.26 \\
MonoFlex~\cite{zhang2021monoflex} &CVPR21 &None &19.94 &13.89 &12.07 &28.23 &19.75 &16.89\\
MonoRCNN~\cite{shi2021geometry} &ICCV21 &None &18.36 &12.65 &10.03  &25.48 &18.11 &14.10 \\
GUPNet~\cite{lu2021GUPNet} &ICCV21 &None &20.11 &14.20 &11.77  &30.29 &21.19 &18.20\\
GAC~\cite{liu2021ground} &RA-L21 &None &21.65 &13.25 &9.91 &29.81 &17.98 &13.08 \\
DeepLine~\cite{liu2021deep} &BMVC21 &None &\textbf{24.23} &14.33 &10.30 &31.09 &19.05 &14.13\\
MonoGround~\cite{qin2022monoground} &CVPR22 &None &21.37 &14.36 &12.62 &30.07 &20.47 &17.74\\
HomoLoss~\cite{gu2022homography} &CVPR22 &None &21.75 &14.94 &13.07 &29.60 &20.68 &17.81\\
DCD~\cite{li2022densely} &ECCV22 &None &23.81 &15.90 &13.21 &\textbf{32.55} &21.50 &18.25\\
\midrule
\textbf{Ours} &- &None &22.43 &\textbf{16.21} &\textbf{13.73} &30.54 &\textbf{21.66} &\textbf{18.64} \\
\bottomrule

\end{tabular}
\end{table*}

\begin{table*}[t]
\small
\setlength{\tabcolsep}{4mm}
\centering
\caption{The performance of $AP_{3D|R_{40}}$ for \textbf{Pedestrian} and \textbf{Cyclist} on the KITTI \emph{\textbf{test}} set.}
\vspace{-10pt}
\label{tab:ped_cyc_result}

\begin{tabular}{ l c c ccc ccc }
\toprule
\multirow{2}{*}{Methods} &\multirow{2}{*}{Year} &\multicolumn{3}{c}{Pedestrian} &\multicolumn{3}{c}{Cyclist} \\
\cmidrule(r){3-8}
& &Easy &Moderate &Hard &Easy &Moderate &Hard \\

\midrule
M3D-RPN~\cite{brazil2019m3d} &ICCV19  &4.92 &3.48 &2.94 &0.94 &0.65 &0.47\\
MonoPair~\cite{chen2020monopair} &CVPR20 &10.02 &6.68 &5.53 &3.79 &2.12 &1.83\\
MonoDLE~\cite{ma2021delving} &CVPR21  &9.64 &6.55 &5.44 &4.59 &2.66 &2.45\\
MonoFlex~\cite{zhang2021monoflex} &CVPR21 &9.43 &6.31 &5.26 &4.17 &2.35 &2.04\\
MonoGround~\cite{qin2022monoground} &CVPR22 &12.37 &7.89 &7.13 &4.62 &2.68 &2.53\\
HomoLoss~\cite{gu2022homography} &CVPR22 &11.87 &7.66 &6.82 &5.48 &3.50 &2.99\\
DCD~\cite{li2022densely} &ECCV22 &10.37 &6.73 &6.28 &4.72 &2.74 &2.41\\
\midrule
\textbf{Ours} &- &11.68 &7.58 &6.22 &5.25 &2.83 &2.31\\
\bottomrule

\end{tabular}
\vspace{-10pt}
\end{table*}

\section{Experiments}

\subsection{Dataset and Metrics}
We first introduce the dataset and evaluation metrics.

KITTI~\cite{Geiger2012CVPR_KITTI}, containing 7,481 training samples and 7,518 testing samples (\emph{test} set), is one of the most widely used datasets for autonomous driving-related tasks. 
We split the training samples into two parts: 3,712 samples for model training (\emph{train} set) and 3,769 for validation (\emph{val} set), following~\cite{ma2021delving}. We test our method on the official server for fair comparisons with other methods.

For evaluation metrics, we report $AP_{3D | R40}$ and $AP_{BEV | R40}$ on car, pedestrian and cyclist categories with detection IoU thresholds of 0.7, 0.5 and 0.5, respectively. For better analysis, all experiments are done under all three difficulties (Easy, Moderate, and Hard) according to 2D bounding box height, occlusion, and truncation degrees.

\subsection{Implementation Details}
 We then explain the details of model hyperparameters and training settings in this section. 

\textbf{Model settings.} Similar to~\cite{ma2021delving}, the input image is first resized to (384, 1280) after random crop and random flip as data augmentation. DLA-34~\cite{yu2018deep} without deformable convolutions is adopted as our backbone network. The feature maps are downsampled by 4$\times$, and the channel of feature maps is 64. The number of query embeddings in Cross Attention is equivalent to the feature map width, i.e., 96. We encode the image features and the fixed position encoding with 1$\times$1 convolution, ReLU, and 1$\times$1 convolution as the input keys in attention. Each head branch comprises 3$\times$3 convolution, ReLU, and 1$\times$1 convolution.

\textbf{Training and Inference}. We implement our method using PyTorch~\cite{paszke2019pytorch}. Our model is trained on two NVIDIA 3090 Ti GPUs in an end-to-end manner for 140 epochs, and the batch size is set to 16. Following~\cite{ma2021delving}, we employ the common Adam optimizer with an initial learning rate of 1.25e-4 and decay it by ten times at 90 and 120 epochs. We also applied the warm-up strategy (5 epochs) to stabilize the training process. In the inference stage, NMS with an IoU threshold of 0.2 is used to filter proposals.

\begin{table}[t]
\small
\setlength{\tabcolsep}{1mm}
\centering
\caption{Comparison of Position-aware methods on KITTI \emph{\textbf{val}} set. We report $AP_{3D|R_{40}}$ and $AP_{BEV|R_{40}}$ for \textbf{Car} category. }
\vspace{-10pt}
\label{tab:ablation_position_aware}

\begin{tabular}{c ccc ccc}
\toprule
\multirow{2}{*}{Methods} &\multicolumn{3}{c}{$AP_{3D|R_{40}}$}    &\multicolumn{3}{c}{$AP_{BEV|R_{40}}$}\\
\cmidrule{2-7}
  &Easy &Mod &Hard &Easy &Mod &Hard \\
\midrule
Baseline &18.3 &14.49 &12.12 &26.11 &20.75 &17.95\\
+CoordConv~\cite{liu2018coordconv} &17.94 &14.76 &12.59 &23.96 &19.63 &17.91\\
+GAC~\cite{liu2021ground} &19.64 &15.20 &12.69 &\textbf{27.64} &21.45 &18.44 \\
+YOLOBU (Ours) &\textbf{20.67} &\textbf{15.81} &\textbf{14.07} &27.53 &\textbf{22.07} &\textbf{19.30}\\
\bottomrule
\end{tabular}
\vspace{-10pt}
\end{table}

\subsection{Comparison with State-of-the-Arts Methods}
In this part, we compare our method with some well-known and effective methods on KITTI~\cite{Geiger2012CVPR_KITTI} dataset.

The results for Car on \emph{test} set are shown in Tab.~\ref{tab:main_result}. We can see that our method outperforms all listed pure image-based methods on the $AP_{3D|R_{40}}$ metric at Moderate and Hard difficulties. 
Specifically, our method outperforms DCD~\cite{li2022densely} by 0.31\%, 0.52\% and HomoLoss~\cite{gu2022homography} by 1.27\%, 0.66\% at Moderate, Hard difficulties respectively. Besides, our method outperforms MonoGround~\cite{qin2022monoground} by 1.06\%, 1.85\%, 1.11\% at Easy, Moderate and Hard settings separately. Although our method is inferior to DeepLine~\cite{liu2021deep} under the Easy setting, it performs better under the Moderate and Hard settings, exceeding 1.88\% and 3.43\%, respectively. Even when compared with listed methods using extra data~\cite{guizilini20203d}, our method is still competitive. Specifically, our method outperforms the knowledge distillation methods SGM3D~\cite{zhou2022sgm3d} by 1.56\% and 0.76\% and MonoDistill~\cite{chong2021monodistill} by 0.18\% and 0.13\% at Moderate and Hard settings. Furthermore, we only get a 0.13\% performance drop at the Moderate setting on $AP_{3D|R_{40}}$ metric when compared with DD3D~\cite{park2021pseudo} that uses a large amount of data in pretraining. We also get the competitive result to DID-M3D~\cite{peng2022did} that leverages Lidar supervision at Moderate and Hard difficulties. This is mainly because our proposed YOLOBU can leverage position clues as supplementary information to locate objects.

The results for Pedestrian and Cyclist on \emph{test} set are shown in Tab.~\ref{tab:ped_cyc_result}.
Although the performance of our method on Pedestrian and Cyclist is not as impressive as the performance on Car, YOLOBU still ranks third and second on Pedestrian and Cyclist, lagging behind the best methods by 0.31\% and 0.67\% at Moderate setting, respectively. 

\subsection{Comparison with Position-Aware Methods}
Our method is significantly better in performance when compared with other position-aware methods~\cite{liu2018coordconv,liu2021ground}. We conduct comparative experiments between CoordConv~\cite{liu2018coordconv}, GAC~\cite{liu2021ground} and our YOLOBU on \emph{val} set. 
For CoordConv~\cite{liu2018coordconv} setting, we simply concatenate the pixels' $u-v$ coordinates and features and then use a 1$\times$1 convolution to extract information. 
For GAC~\cite{liu2021ground} setting, we use the ground-aware convolution in regression heads following the implementation of its official codebase.
The results are listed in Tab.~\ref{tab:ablation_position_aware}, which shows that our YOLOBU outperforms the other two methods by a large margin on $AP_{3d}$. Specifically, YOLOBU beats CoordConv ~\cite{liu2018coordconv} by 2.73\%, 1.05\%, 1.48\% on $AP_{3d}$ at Easy, Moderate, Hard difficulty respectively. YOLOBU also exceeds GAC~\cite{liu2021ground} by 1.03\%, 0.61\%, 1.38\% on $AP_{3D}$ at Easy, Moderate, Hard difficulty respectively. This is mainly because our YOLOBU models the bottom-up relationships between pixels and thus encodes more representative position information for monocular 3D detection.

\subsection{Ablation Study}
We further conduct additional experiments to analyze the effectiveness of each component. We use MonoDLE~\cite{ma2021delving} as our baseline and keep all experimental settings the same. 

\begin{table}[t]
\small
\setlength{\tabcolsep}{0.8mm}
\centering
\caption{Ablation study of core modules on KITTI \emph{\textbf{val}} set. We report $AP_{3D|R_{40}}$ and $AP_{BEV|R_{40}}$ for \textbf{Car}. }
\vspace{-10pt}
\label{tab:ablation}

\begin{tabular}{c cc ccc ccc}
\toprule
\multirow{2}{*}{Methods} &\multirow{2}{*}{CCA} &\multirow{2}{*}{RRCS}  &\multicolumn{3}{c}{$AP_{3D|R_{40}}$} &\multicolumn{3}{c}{$AP_{BEV|R_{40}}$}\\
\cmidrule{4-9}
 & & &Easy &Mod &Hard &Easy &Mod &Hard \\
\midrule
Baseline & & &18.3 &14.49 &12.12 &26.11 &20.75 &17.95\\
(a) &$\checkmark$ & &19.07 &14.87 &12.67 &26.05 &19.92 &17.28\\
(b) & &$\checkmark$ &17.43 &13.67 &12.19 &24.58 &19.37 &16.84\\
(c) &$\checkmark$ &$\checkmark$ &\textbf{20.67} &\textbf{15.81} &\textbf{14.07} &\textbf{27.53} &\textbf{22.07} &\textbf{19.30}\\
\bottomrule
\end{tabular}
\vspace{-10pt}
\end{table}

\textbf{Effectiveness of the CCA.} To figure out how much contribution the Column-based Cross Attention (CCA) brings, we attach the CCA to the backbone of our baseline. As Tab.~\ref{tab:ablation}(a) shows, the CCA brings $AP_{3D}$ (0.77\%, 0.38\% and 0.55\% improvements on Easy, Moderate and Hard difficulties, respectively) while slightly reducing in the $AP_{BEV}$ metric. The results indicate that the contribution of the CCA only is limited since \emph{the positional relation between pixels is not modeled}.

\textbf{Effectiveness of the RRCS.} 
To validate the effectiveness of Row-based Reverse Cumulative Sum (RRCS) only, we insert the proposed RRCS after the backbone of our baseline. The results are listed in Tab.~\ref{tab:ablation}(b). We can observe that only adding RRCS is detrimental to the performance. Note that the semantic and positional information of each pixel is variant and keeping all the edge weights the same would bring much noise to the model.

We emphasize that \emph{the CCA and RRCS are an integral whole to establish the association of pixels}. The former generates the edge values, and the latter connects nodes with the edge values. As shown in Tab.~\ref{tab:ablation}(c), when we use both the CCA and RRCS, we can obtain a significant improvement over the baseline. Our method brings 2.37\%, 1.32\%, 1.95\% improvement of $AP_{3D}$ on Easy, Moderate, and Hard difficulties respectively, demonstrating the significance of the positional relation modeling.

\textbf{Analysis of attention types.} Tab.~\ref{tab:ablation_attention} quantifies the influence of different attention types in the proposed CCA. For the model without column-based attention, we use a single query for all pixels and conduct global cross attention instead.  We can observe that column-based attention is necessary since the semantics of structured layout pixels (e.g., road pixels) vary across columns, and one single query is not representative enough. 

\begin{table}[t]
\vspace{10pt}
\footnotesize
\setlength{\tabcolsep}{1.25mm}
\centering
\caption{Ablation study of attention methods in CCA on KITTI \emph{\textbf{val}} set. We report $AP_{3D|R_{40}}$ and $AP_{BEV|R_{40}}$ for \textbf{Car}.}
\vspace{-10pt}
\label{tab:ablation_attention}

\begin{tabular}{c ccc ccc}

\toprule
\multirow{2}{*}{Methods} &\multicolumn{3}{c}{$AP_{3D|R_{40}}$}    &\multicolumn{3}{c}{$AP_{BEV|R_{40}}$}\\
\cmidrule{2-7}
  &Easy &Mod &Hard &Easy &Mod &Hard \\
\midrule
w/o column-based &17.95 &14.77 &12.63 &25.85 &21.08 &18.39\\
w column-based &\textbf{20.67} &\textbf{15.81} &\textbf{14.07} &\textbf{27.53} &\textbf{22.07} &\textbf{19.30}\\
\bottomrule
\end{tabular}
\vspace{-5pt}
\end{table}

\begin{table}[t]
\footnotesize
\setlength{\tabcolsep}{2mm}
\centering
\caption{Ablation study of Cumulative Sum direction in RRCS on KITTI \emph{\textbf{val}} set. We report $AP_{3D|R_{40}}$ and $AP_{BEV|R_{40}}$ for \textbf{Car}.}
\vspace{-10pt}
\label{tab:ablation_cumsum}

\begin{tabular}{c ccc ccc}
\toprule
\multirow{2}{*}{Methods} &\multicolumn{3}{c}{$AP_{3D|R_{40}}$}    &\multicolumn{3}{c}{$AP_{BEV|R_{40}}$}\\
\cmidrule{2-7}
  &Easy &Mod &Hard &Easy &Mod &Hard \\
\midrule
Baseline &18.3 &14.49 &12.12 &26.11 &20.75 &17.95\\
Up-Bottom &19.37 &14.79 &13.33 &26.34 &20.95 &18.40\\
Bottom-Up &\textbf{20.67} &\textbf{15.81} &\textbf{14.07} &\textbf{27.53} &\textbf{22.07} &\textbf{19.30}\\
\bottomrule
\end{tabular}
\vspace{-10pt}
\end{table}

\textbf{Analysis of the Cumulative Sum direction.} In order to examine the impact of different cumulative sum directions (i.e., Bottom-Up and Up-Bottom), we conduct comparative experiments and the results are listed in Tab.~\ref{tab:ablation_cumsum}. We can observe that building the graph either in the Bottom-Up or Up-Bottom direction improves the performance over baseline, which proves the effectiveness of pixels' association in columns for the monocular 3D detection task. Meanwhile, we can see that building relations from the Bottom-Up direction is much better than Up-Bottom. This is consistent with our prior that the pixels which help locate objects mainly occur below them.

\begin{figure}[t]
\vspace{10pt}
\centering
\includegraphics[width=0.97\linewidth]{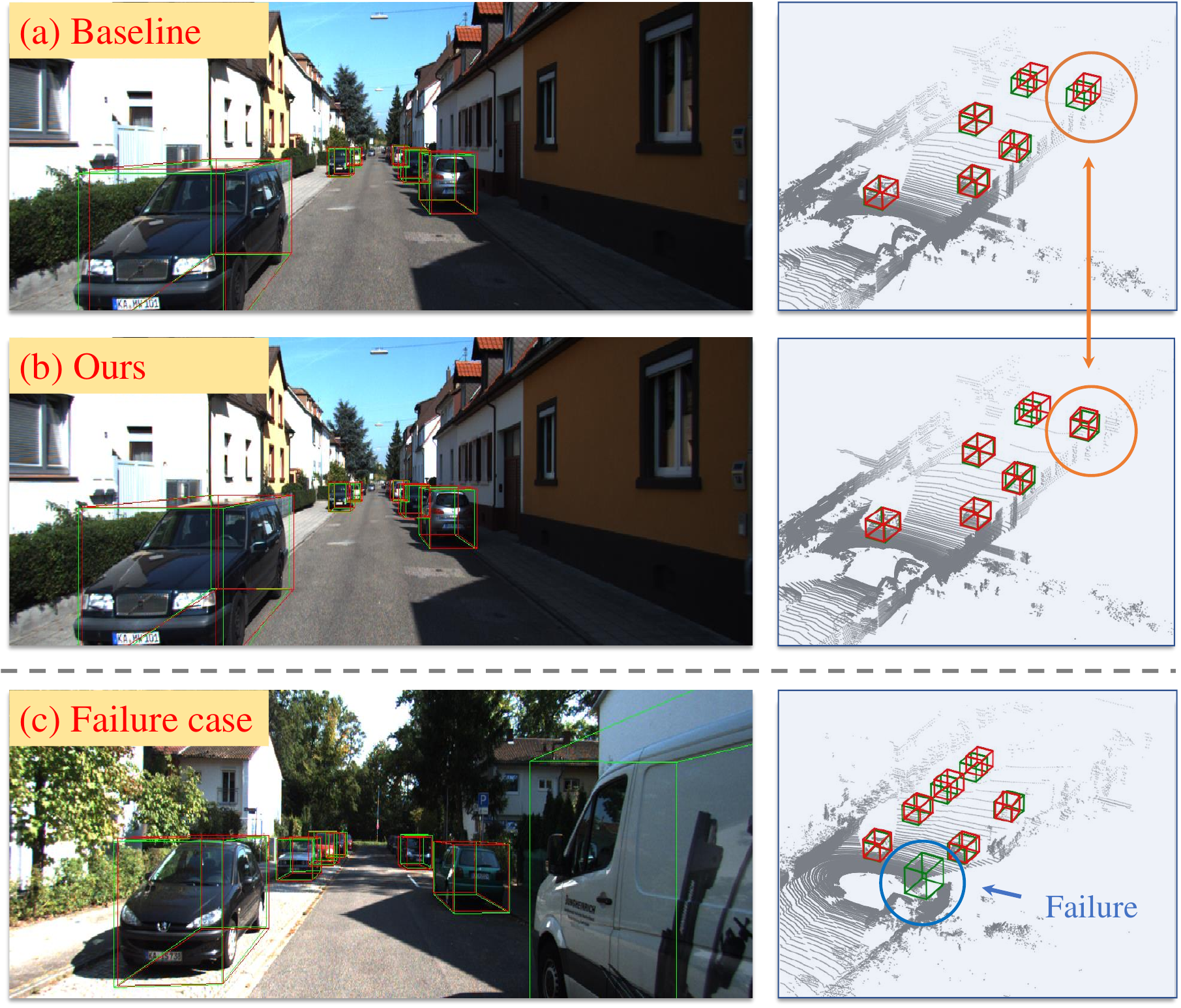}
\caption{Visualizations on KITTI \emph{val} set. (a) Baseline (MonoDLE~\cite{ma2021delving}); (b) The proposed YOLOBU; (c) Failure case from YOLOBU. \textcolor{red}{Red} and \textcolor{green}{green} bounding boxes indicate predictions and ground truth, respectively. The LiDAR point clouds are only used for visualization.}
\centering
\label{fig:vis}
\vspace{-10pt}
\end{figure}

\begin{table}[t]
\small
\setlength{\tabcolsep}{1mm}
\centering
\caption{The performance on the nuScenes \emph{\textbf{val}} set. $^*$ denotes that the results is reproduced using the official code base without the finetune stage.}
\label{tab:main_nuscenes}

\begin{tabular}{ l | c c | c c c c c }
\toprule
Methods & NDS & mAP & mATE & mASE & mAOE & mAVE & mAAE\\
\midrule
FCOS3D$^*$& 0.365 & 0.287 & 0.814 & 0.269 & 0.530 & 1.319 & 0.171 \\ 
\textbf{Ours} & \textbf{0.381} & \textbf{0.303} & \textbf{0.762} & \textbf{0.268} & \textbf{0.515} & \textbf{1.316} & \textbf{0.158}\\
\bottomrule

\end{tabular}
\end{table}

\subsection{Performance on the Large-scale nuScenes Dataset}
To further evaluate the effectiveness of our method, we conduct a comparison experiment on the large-scale nuScenes dataset. We train models on the official train split and evaluate on the val split. For evaluation metrics, we report the official metrics include: Nuscenes Detection Score (NDS), mean Average Precision (mAP),  Average Translation Error (ATE), Average Scale Error (ASE), Average Orientation Error (AOE), Average Velocity Error (AVE) and Average Attribute Error (AAE).

We use the FCOS3D~\cite{wang2021fcos3d} as the baseline, and we insert the proposed modules (CCA and RRCS) after the neck to refine the features of each scale. Because of the significant increase in the scale of training data, we use 8 NVIDIA 3090 GPUs with a total batch of 32 to train models. The results are listed in Tab.~\ref{tab:main_nuscenes}. Our method outperforms the FCOS3D by 1.6\% NDS and mAP, showing its effectiveness, and the conclusions are consistent with that of the KITTI dataset.

\subsection{Visualizations and Discussions}

We visualize the model predictions to analyze our method more intuitively. The results are shown in Fig.\ref{fig:vis}.

\textbf{Qualitative results.} As we can see, our proposed YOLOBU predicts remarkably precise 3D Bounding Boxes in scenario (b). Specifically, compared with our baseline MonoDLE~\cite{ma2021delving} in scenario (a), our method can correctly detect the car in the top right corner of the image while the baseline predicts an inaccurate bounding box. This phenomenon demonstrates that image-based 3D detection methods benefit from the bottom-up positional information.

\textbf{Generalization for other applications.} Since our method effectively leverages the position clues from images and serves as a plug-and-play module, one can easily adapt our method for other applications by simply plugging it into the original solution, as long as the task (e.g., stereo and multi-view 3D object detection for research, and autonomous driving, underwater robots and intelligent logistics for the industry) meets the assumptions introduced by our method.

\textbf{Limitations.} Our method has some limitations: 1) Our method fails to detect the truncated and large objects, as shown in Fig.~\ref{fig:vis} (c). The main reason is that these samples cannot benefit from the bottom-up positional information. 2) The inference time of the proposed method will be slightly slower than the baseline~\cite{ma2021delving} (22.56 FPS vs. 23.64 FPS). We will explore real-time solutions in the future. 
3) Our method does not model the calibration parameters, and the performance may be affected when the calibration parameter change. However, we can easily modify the detection head to model them explicitly for better results.

\section{CONCLUSIONS}
In this paper, we demonstrate that intrinsic position clues from images are important but ignored by most existing image-based methods. We propose a novel method for Monocular 3D Object Detection named YOLOBU, which performs cross attention on each column and conduct cumulative sum in the bottom-up direction in a row-by-row way. With these two steps, our method is able to learn position information from bottom to up and serve as strong auxiliary clues to solve the size ambiguity problem. Our method has achieved competitive performance on the KITTI 3D object detection benchmark using a monocular camera without additional information. We hope our neat and effective approach will serve as a strong baseline for future research in Monocular 3D Object Detection.


%




\ifCLASSOPTIONcaptionsoff
  \newpage
\fi



\bibliographystyle{IEEEtran}
\bibliography{reference}

%



%




\end{document}